\documentclass[runningheads]{llncs}

\usepackage{graphicx}
\usepackage{amsmath}
\usepackage{float}
\usepackage{a4wide}
\usepackage{algorithm}
\usepackage{algorithmicx}
\usepackage{algpseudocode}
\usepackage{epstopdf}
\usepackage{bm}
\usepackage{tikz}
\usepackage{multirow}
\usepackage{amsmath}
\usetikzlibrary{bayesnet}
\usetikzlibrary{shapes.gates.logic.US,trees,positioning,arrows}
\usepackage{caption}
\usepackage{subcaption}
\usetikzlibrary{trees}
\usepackage{mathtools}
\usepackage{amsmath}
\usepackage{booktabs}
\usepackage{rotating}

\usepackage{epstopdf}
\usepackage{amssymb}
\usepackage{microtype}
\usepackage{url}
\usepackage{caption}
\usepackage{subcaption}

\usepackage[T1]{fontenc}
\usepackage{lmodern}

\begin{document}

	\title{Mitigating sampling bias in risk-based active learning via an EM algorithm}
	
	%
	%
	\author{A.J.\ Hughes\inst{1} \and L.A.\ Bull\inst{2} \and
		P.\ Gardner\inst{1} \and N.\ Dervilis \inst{1} \and K.\ Worden\inst{1}}
	\authorrunning{A.J.\ Hughes et al.}
	%
	\institute{Dynamics Research Group, Department of Mechanical Engineering, University of Sheffield, Sheffield, \\ S1 3JD, UK
		 \and
		The Alan Turing Institute, The British Library, 96 Euston Road, London, NW1 2DB, UK\\
	\email{aidan.j.hughes@sheffield.ac.uk}}
	\maketitle              
	
	\begin{abstract} 
	
	Risk-based active learning is an approach to developing statistical classifiers for online decision-support. In this approach, data-label querying is guided according to the expected value of perfect information for incipient data points. For SHM applications, the value of information is evaluated with respect to a maintenance decision process, and the data-label querying corresponds to the inspection of a structure to determine its health state. Sampling bias is a known issue within active-learning paradigms; this occurs when an active learning process over- or undersamples specific regions of a feature-space, thereby resulting in a training set that is not representative of the underlying distribution. This bias ultimately degrades decision-making performance, and as a consequence, results in unnecessary costs incurred.
	
	The current paper outlines a risk-based approach to active learning that utilises a semi-supervised Gaussian mixture model. The semi-supervised approach counteracts sampling bias by incorporating pseudo-labels for unlabelled data via an EM algorithm. The approach is demonstrated on a numerical example representative of the decision processes found in SHM.

	\keywords{active learning \and semi-supervised learning \and decision-making \and risk \and value of information}
	\end{abstract}

	\section{Introduction}  

	In structural health monitoring (SHM) systems, statistical pattern recognition (SPR), can be used to classify feature vectors $\mathbf{x} \in \mathbb{R}^D$ extracted from data, according to a descriptive label $y \in \{ 1,\ldots,K\}$, that may correspond to damage location, type or extent \cite{Farrar2013}. To account for the inherent uncertainty associated with monitoring structures, statistical classifiers are used to infer probability distributions over health states, i.e.\ $p(y=k|\mathbf{x})$. Gaining the ability to make robust and informed decisions regarding the operation and maintenance of structures provides significant motivation for the development and implementation of SHM technologies.
	
	In \cite{Hughes2022}, a method is proposed for considering the decision-support application within the learning process for the statistical classifiers used in SHM. Termed \textit{risk-based active learning}, the approach utilises \textit{expected value of perfect information} (EVPI) to trigger structural inspections, in order to obtain health-state labels for observations. Using these labelled data, a classification model is subsequently relearned/updated. In \cite{Hughes2022}, it is shown that decision-making performance is improved via risk-based active learning, however, a slight decline in decision-making performance is observed during the later stages of the risk-based active learning process. This degradation in performance can be attributed to \textit{sampling bias} \cite{Dasgupta2008,Dasgupta2011}.
	
	Sampling bias occurs due to the fact that risk-based active learning preferentially queries data with high expected value of information. The preferential querying process results in specific regions of the feature space and certain classes being disproportionately represented in the training data. This bias is particularly problematic for generative classification models as the learning process typically relies on the assumption that training data are representative of the underlying distribution. The bias introduced via the queried samples facilitates the learning of decision boundaries in a cost-effective manner; however, it also causes a decline in decision-making performance for later queries.
	
	The current paper aims to mitigate the effects of sampling bias via introducing semi-supervised learning into the risk-based active learning approach. Semi-supervised methods incorporate unlabelled data into the learning process, thereby rectifying the bias in the training dataset introduced by active learning. It is demonstrated here, via numerical example, that semi-supervised learning (specifically, the expectation-maximisation algorithm) can improve risk-based active learning by reducing the effects of sampling bias; improving decision-making performance, and reducing the number of inspections made throughout a monitoring campaign. The layout of the paper is as follows. Firstly, background information on risk-based active learning and semi-supervised learning are provided. Subsequently, risk-based active learning with semi-supervised learning incorporated is demonstrated on a synthetic dataset. Finally, concluding remarks are provided.

	\section{Risk-based Active Learning}
	
	Active learning is a form of partially-supervised learning. Partially-supervised learning is characterised by the use of both labelled and unlabelled data.
	
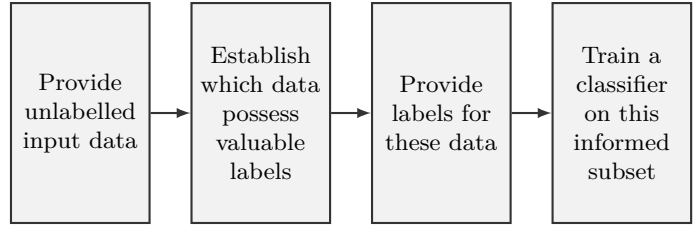
\begin{figure}[ht!]
		\centering
		\begin{tikzpicture}[auto]
			\begin{footnotesize}
				\tikzstyle{block} = [rectangle, thick, draw=black!80, text width=5em, text centered, minimum height=9em, fill=black!5]
				\tikzstyle{line} = [draw, -latex, thick]
				\node [block, node distance=24mm] (A) {Provide\\ unlabelled input data};
				\node [block, right of=A, node distance=24mm] (B) {Establish which data possess valuable labels};
				\node [block, right of=B, node distance=24mm] (C) {Provide labels for these data};
				\node [block, right of=C, node distance=24mm] (D) {Train a classifier on this informed subset};
				\path [line, draw=black!80] (A) -- (B);
				\path [line, draw=black!80] (B) -- (C);
				\path [line, draw=black!80] (C) -- (D);
			\end{footnotesize}
		\end{tikzpicture}
		\caption{A general active learning heuristic.}
		\label{fig:AL1}
	\end{figure}

	Active-learning algorithms construct a labelled data set $\mathcal{D}_l$ by querying labels for otherwise unlabelled data $D_u$. Labels for incipient data are queried preferentially according to some measure of their desirability. Models are then trained in a supervised manner using $\mathcal{D}_l$. A general heuristic for active learning is presented in Figure \ref{fig:AL1}.
	
	In \cite{Bull2019}, an online active learning-approach is shown to overcome the primary challenge for classifier development in the context of SHM - initial scarcity of comprehensive labelled data. This result is achieved via the construction of a labelled dataset from a querying process that corresponds to the inspection of a structure by an engineer to determine its health state. In \cite{Hughes2022}, an alternative risk-based approach to active learning is proposed in which \textit{expected value of perfect information} (EVPI) is used as a measure to guide querying.
	
	The expected value of perfect information (EVPI) can be interpreted as the price that a decision-maker should be willing to pay in order to gain access to perfect information of an otherwise unknown or uncertain state. Here, it should be clarified that the terminology `perfect information' refers to ground-truth states observed without uncertainty. More formally, the expected value of observing $y$ with perfect information prior to making decision $d$ is given by,
	
	\begin{equation}\label{eq:EVPI}
		\text{EVPI}(d | y) := \text{MEU}(\mathcal{I}_{y \rightarrow d}) - \text{MEU}(\mathcal{I})
	\end{equation}
	
	\noindent
	where $\mathcal{I}$ is an influence diagram representing an SHM decision process with decision $d$ influenced by random variable $y$, and $\mathcal{I}_{y \rightarrow d}$ is a modified influence diagram incorporating an additional informational link from $y$ to $d$ indicating that $y$ is observed prior to $d$. An example calculation of EVPI is presented in \cite{Hughes2022}. Details of the procedure for forming SHM decision processes as influence diagrams can be found in \cite{Hughes2021}.
	
	A convenient criterion for mandating structural inspections can be derived from the EVPI. Put simply, if the EVPI of a label $y_t$ for a data point $\tilde{\mathbf{x}}_t$ exceeds the cost of making an inspection $C_{\text{ins}}$ then the ground-truth for $y_t$ should be obtained prior to $d_t$. Subsequently, the labelled dataset $\mathcal{D}_l$ can be extended to include the newly-acquired data-label pair $(\mathbf{x}_t,y_t)$, and the classifier retrained. While the assumption of perfect information may not hold in all cases, the principles and methodologies discussed in the current paper hold in general for value of information. Furthermore, the perfect information assumption may be relaxed via the introduction of an additional probabilistic model that quantifies uncertainty in inspections. Full details of the risk-based active-learning algorithm can be found in \cite{Hughes2022}.
	
	Adopting a risk-based approach to active learning allows one to learn statistical classifiers with consideration for the decision support contexts in which they may be employed. In \cite{Hughes2022}, it is demonstrated that this approach provides a cost-efficient manner for classifier development.

	\section{Semi-supervised Learning}
	
	Alongside active learning, semi-supervised learning is a form of partially-supervised learning -- utilising both $\mathcal{D}_l$ and $\mathcal{D}_u$ to inform the classification mapping. The fundamental principle of semi-supervised learning that distinguishes it from active learning, is as follows; data in the unlabelled dataset $\mathcal{D}_u$ can be given \textit{pseudo-labels} that are informed by the ground-truth labels available in $\mathcal{D}_l$. By incorporating pseudo-labels for unqueried data into the risk-based active-learning algorithm, class imbalance and inadequate coverage of the feature space can be rectified.
	
	Semi-supervised learning has been applied to pattern recognition problems for SHM \cite{Bull2020}. This method of learning brings several benefits such as increased the utilisation of information obtained via costly structural inspections.
	
	\subsection{Expectation-Maximisation}
	
	The approach to semi-supervised learning considered in the current paper aims to exploit a generative model form of statistical classifier, specifically a Gaussian mixture model, the details of which can be found in \cite{Bull2019,Hughes2022}. Generative models can conveniently account for labelled and unlabelled data. This capability is achieved by modifying the \textit{expectation-maximisation} (EM) algorithm \cite{Dempster1977}, typically used for unsupervised density estimation, such that the log-likelihood of the model is maximised over both unlabelled and labelled data. For details of the approach, the reader is referred to \cite{Bull2020}.
	
	Within the risk-based active learning algorithm, it is possible to apply EM every time a new unlabelled observation is acquired; however, this would be computationally expensive. To limit the computational cost of the modified active learning algorithm, the EM update was only applied following the retraining of the model subsequent to the acquisition of a new ground-truth label obtained via inspection.

	\section{Case Study}
	
	To draw attention to the effects of sampling bias in risk-based active learning algorithms, a modified version of the synthetic dataset used in \cite{Hughes2022} is adopted. The dataset consists of a two-dimensional input space $\mathbf{x}_t = \{ x_t^{1} , x_t^{2} \}$ and a four-class label space $y_t \in \{ 1,2,3,4 \}$. Framed as an SHM dataset, one can consider the inputs as features used to discriminate between classes corresponding to four health states of interest for a structure $S$ where:
	
	\begin{itemize}
		\item Class 1 corresponds to the structure being undamaged and fully functional.
		\item Class 2 corresponds to the structure possessing minor damage whilst remaining fully functional.
		\item Class 3 corresponds to the structure possessing significant damage resulting in a reduced operational capacity.
		\item Class 4 corresponds to the structure possessing critical damage resulting in operational incapacity, i.e.\ total structural failure.
	\end{itemize}
	
\begin{figure}[ht!]
		\begin{subfigure}{.5\textwidth}
			\centering
			\scalebox{0.4}{
				\includegraphics{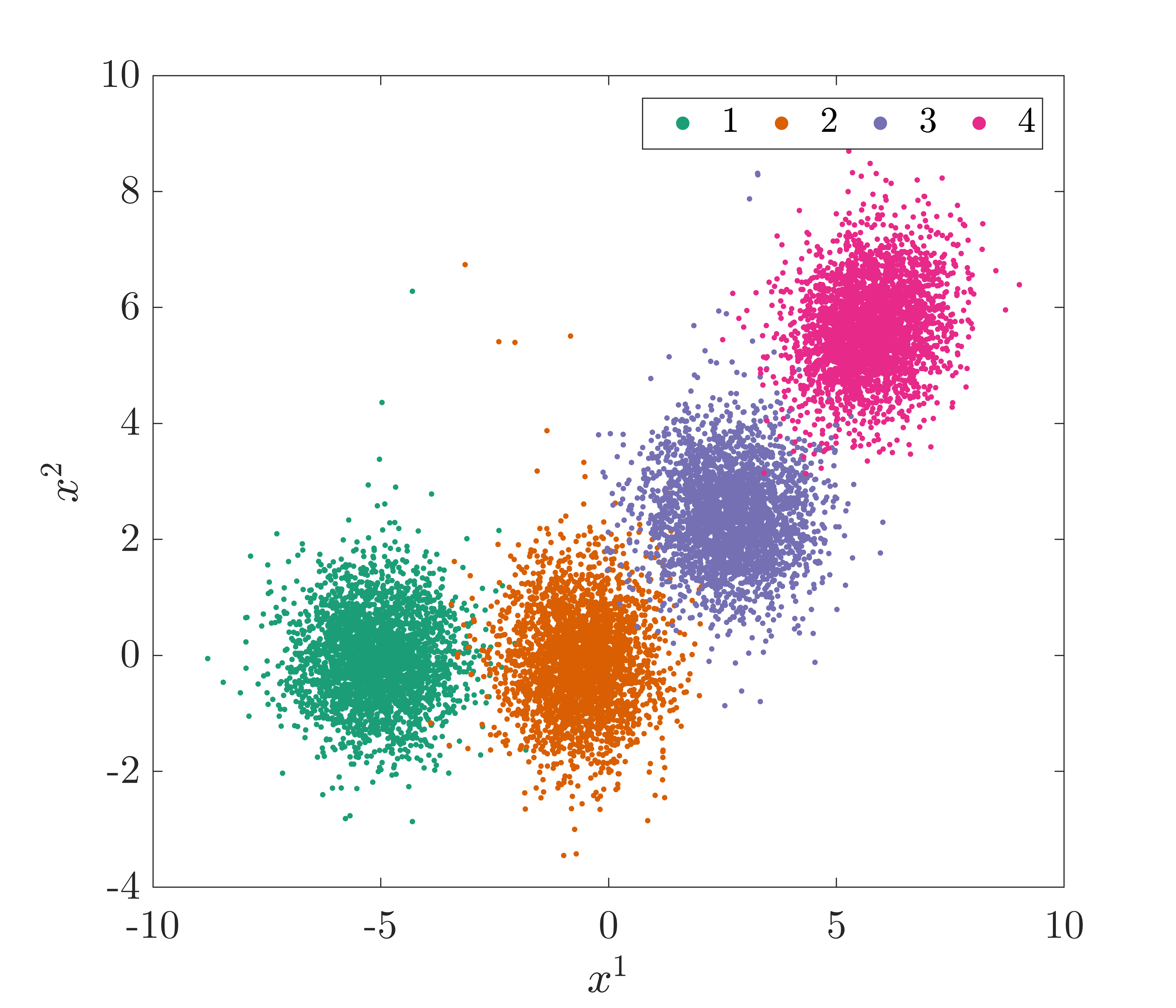}
			}
			\caption{ }
			\label{fig:scatter_ext}
		\end{subfigure}
		\begin{subfigure}{.5\textwidth}
			\centering
			\scalebox{0.4}{
				\includegraphics{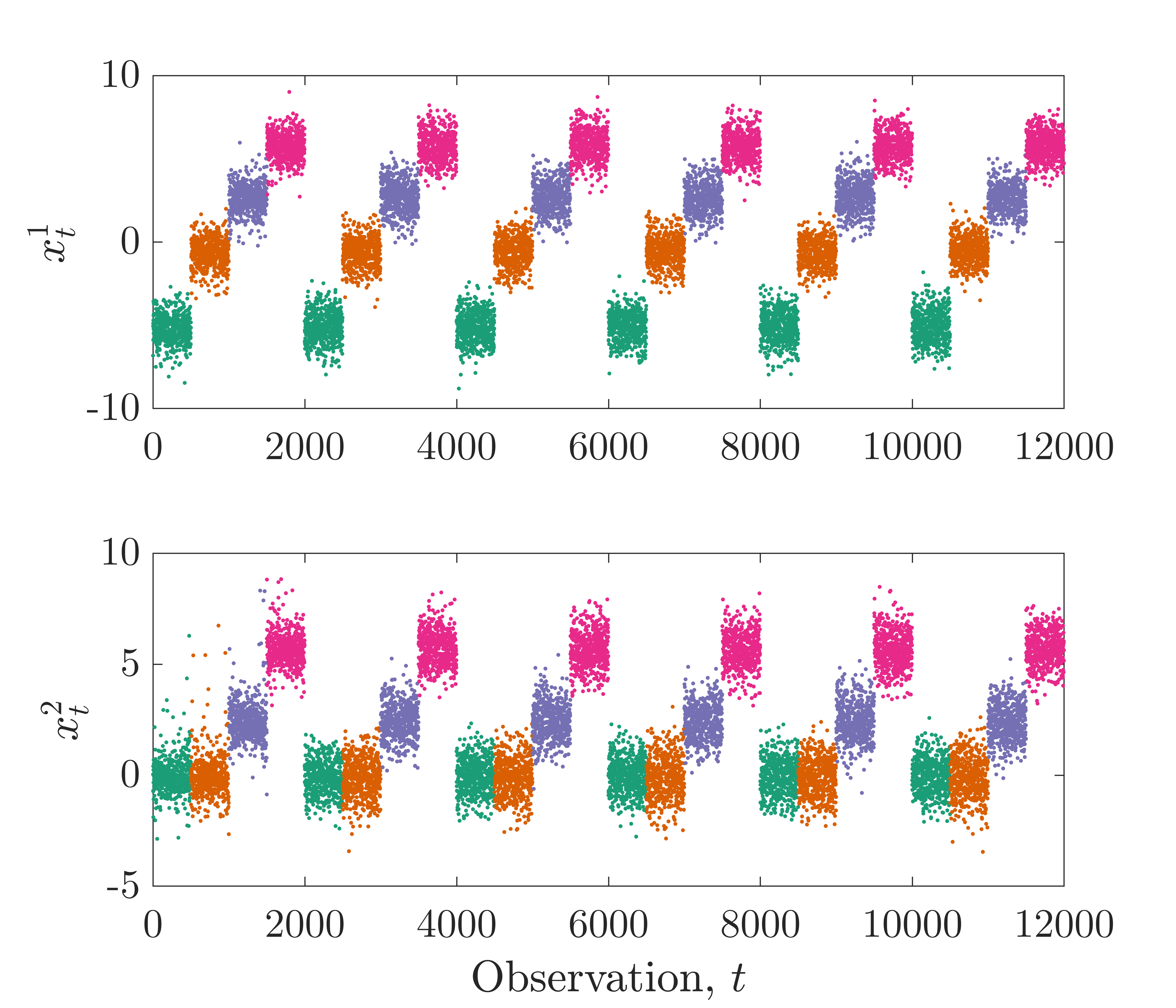}
			}
			\caption{ }
			\label{fig:overview_ext}
		\end{subfigure}
		\caption{Visualisation of the extended synthetic dataset in (a) the feature space and (b) discrete time $t$.}
		\label{fig:data_ext}
	\end{figure}

	The original dataset comprised of 1997 data points ordered according to health state -- from Class 1 through to Class 4. For the current case study, the dataset was extended by drawing an additional 2000 independent samples from a generative distribution learned from the original dataset. This procedure was conducted a total of five times, resulting in 11997 data points that repeatedly progress from Class 1 through to Class 4, thereby emulating the process of structural deterioration and subsequent repair - a pattern that could conceivably be experienced by the fictitious structure of interest $S$. A visualisation of the extended dataset is presented in Figure \ref{fig:data_ext}.
	
	\subsection{Decision Process}
	
	In order to employ risk-based active learning for the development of a statistical classifier, a decision process for the structure $S$ must first be specified. For consistency with \cite{Hughes2022}, an identical binary maintenance decision process is selected for the current study.
	
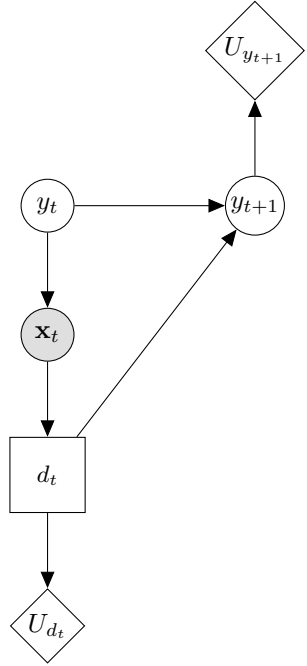
\begin{figure}[ht!]
		\centering
		\begin{tikzpicture}[x=1.7cm,y=1.8cm]
			
			\node[det] (uf2) {$U_{y_{t+1}}$} ;
			\node[latent, below=1cm of uf2] (x2) {$y_{t+1}$} ;
			\node[latent, left=2cm of x2] (x1) {$y_{t}$} ;
			\node[obs, below=1cm of x1] (y1) {$\mathbf{x}_{t}$} ;
			\node[rectangle,draw=black,minimum width=1cm,minimum height=1cm,below=1cm of y1] (d1) {$d_{t}$} ;
			\node[det, below=1cm of d1] (u1) {$U_{d_{t}}$} ;

			\edge {x2} {uf2} ; %
			\edge {x1} {x2} ; %
			\edge {x1} {y1} ; %
			\edge {d1} {x2} ; %
			\edge {d1} {u1} ; %
			\edge {y1} {d1} ; %

		\end{tikzpicture}
		\caption{An influence diagram representation of the decision process associated with structure $S$.}
		\label{fig:ID_ext}
	\end{figure}

	Consider an agent that, at discrete-time instances $t \in \mathbb{N}$, is tasked with making a binary decision $d_t$. The agent must select an action from $\text{dom}(d_t) = \{ 0 \text{ (do nothing)} , 1 \text{ (repair)} \}$, such that some degree of operational capacity is maintained for $S$ at instance $t+1$. It is assumed that the agent has access to the observed discriminative features $\mathbf{x}_t$, and must infer the current and future latent health structural states $y_t$ and $y_{t+1}$ via a statistical classifier and a transition model, respectively. An influence diagram for such a decision process is shown in Figure \ref{fig:ID_ext}.
	
	As aforementioned, the agent requires a transition model in order to obtain a forecast for the future health state $y_{t+1}$. For the current case study, it is assumed that this model is known \textit{a priori} so that focus can be on the development of the statistical classifier. Examples of this transition model being developed from data and prior knowledge of physics are presented in \cite{Vega2020a} and \cite{Hughes2022b}, respectively. As shown in Figure \ref{fig:ID_ext}, the future health state $y_{t+1}$ is conditionally dependent on the current health state $y_{t}$ and the decision $d_t$. Given $d_t = 0$, it is assumed that the structure will monotonically degrade with a propensity to remain in its current health state. This assumption is reflected in the conditional probability distribution $P(y_{t+1}|y_t,d_t=0)$ presented in Table \ref{tab:P_y1_y0_d0}. Given $d_t = 1$, it is assumed that the structure is returned to its undamaged state with probability 0.99 and remains in its current state with probability 0.01. This assumption is reflected in the conditional probability distribution $P(y_{t+1}|y_t,d_t=1)$ presented in Table \ref{tab:P_y1_y0_d1}.
	
	\begin{table}
		\begin{minipage}{.5\linewidth}
			\centering
			\caption{The conditional probability \\ table $P(y_{t+1}|y_t, d_t)$ for $d_t = 0$.}
			\label{tab:P_y1_y0_d0}       
			\begin{tabular}{c c c c c c}
				\toprule
				\midrule
				& & \multicolumn{4}{c}{$y_{t+1}$}\\
				&& 1  & 2 & 3 & 4  \\ \cmidrule{3-6}
				\multicolumn{1}{c}{\multirow{4}{*}{\begin{sideways}\parbox{1.5cm}{\centering $y_t$}\end{sideways}}}   &
				\multicolumn{1}{l}{1}& 0.8 & 0.18 & 0.015 & 0.005 \\
				\multicolumn{1}{c}{}    &
				\multicolumn{1}{l}{2}& 0 & 0.8 & 0.15 & 0.05  \\
				\multicolumn{1}{c}{}    &
				\multicolumn{1}{l}{3} & 0 & 0 & 0.8 & 0.2  \\
				\multicolumn{1}{c}{}    &   
				\multicolumn{1}{l}{4} & 0 & 0 & 0 & 1  \\
				\midrule
				\bottomrule
			\end{tabular}
		\end{minipage}%
		\begin{minipage}{.5\linewidth}
			\centering
			\caption{The conditional probability \\ table $P(y_{t+1}|y_t, d_t)$ for $d_t = 1$.}
			\label{tab:P_y1_y0_d1}       
			\begin{tabular}{c c c c c c}
				\toprule
				\midrule
				& & \multicolumn{4}{c}{$y_{t+1}$}\\
				&& 1  & 2 & 3 & 4  \\ \cmidrule{3-6}
				\multicolumn{1}{c}{\multirow{4}{*}{\begin{sideways}\parbox{1.5cm}{\centering $y_t$}\end{sideways}}}   &
				\multicolumn{1}{l}{1}& 1 & 0 & 0 & 0 \\
				\multicolumn{1}{c}{}    &
				\multicolumn{1}{l}{2}& 0.99 & 0.01 & 0 & 0  \\
				\multicolumn{1}{c}{}    &
				\multicolumn{1}{l}{3} & 0.99 & 0 & 0.01 & 0  \\
				\multicolumn{1}{c}{}    &   
				\multicolumn{1}{l}{4} & 0.99 & 0 & 0 & 0.01  \\
				\midrule
				\bottomrule
			\end{tabular}
		\end{minipage}
	\end{table}

	For simplicity, and for more compact influence diagrams, within the current case study it is assumed that health states may be mapped directly to utilities. Once again, the utility function used here is specified to reflect the relative utility values that may be expected in a typical SHM application. The utility function $U(y_{t+1})$ is provided in Table \ref{tab:Uy}. It can be seen from Table \ref{tab:Uy}, that for health-states 1 and 2, in which the structure is at full operational capacity, positive utility has been assigned. For health-state 3, which corresponds to a reduced operational capacity, a lesser positive utility has been assigned. For health-state 4, which corresponds to critical damage, a relatively-large negative utility has been assigned, to reflect a complete loss of operational capacity and an additional severe consequence associated with structural failure, e.g.\ environmental damage.
	
	Figure \ref{fig:ID_ext} indicates that the actions in the domain of $d_t$ have associated costs, specified by the utility function $U(d_t)$ and shown in Table \ref{tab:Ud}. It is assumed that $d_t = 0$ has no utility associated with it. On the other hand, $d_t = 1$ has negative utility because of the expenditure necessary to cover the material and labour costs associated with structural maintenance. It is worth noting that, in many practical applications, the specification of utility function is non-trivial and is an active topic of research outside the scope of the current paper. Hence, for this case study, relative utility values are selected to be only somewhat representative of the SHM context.

	\begin{table}
		\begin{minipage}{.5\linewidth}
			\centering
			\caption{The utility function $U(y_{t+1})$.}
			\label{tab:Uy}   
			\begin{tabular}{cc}
				\toprule
				\midrule
				$y_{t+1}$ & $U(y_{t+1})$\\
				\midrule
				$1$ & $10$\\
				$2$ & $10$\\
				$3$ & $5$\\
				$4$ & $-75$\\
				\midrule
				\bottomrule
			\end{tabular}
		\end{minipage}%
		\begin{minipage}{.5\linewidth}
			\centering
			\caption{The utility function $U(d_t)$ where $d_t=0$ and $d_t=1$ denote the `do nothing' and `repair' actions, respectively.}
			\label{tab:Ud}   
			\begin{tabular}{cc}
				\toprule
				\midrule
				$d_t$ & $U(d_t)$\\
				\midrule
				$0$ & $0$\\
				$1$ & $-30$\\
				\midrule
				\bottomrule
			\end{tabular}
		\end{minipage}
	\end{table}

	As alluded to previously, a probability distribution over the current health state $y_t$ is inferred from the observed features $\mathbf{x}_t$ via a statistical classifier subject to risk-based active learning -- for the current case study a four-class Gaussian mixture model is used. To fully specify the contextual parameters for risk-based active learning, it is assumed that the ground-truth health state at discrete-time instance $t$ can be obtained via inspection at the cost of $C_{\text{ins}} = 7$. Here, cost can be interpreted as a strictly non-positive utility.

	\subsection{Results}
	
	In order to assess the effects of sampling bias on decision-making performance and for comparison to highlight the effects of introducing semi-supervised learning, risk-based active learning was used to learn a GMM for use within the decision process outlined in the previous section. This process was repeated 1000 times. For each repetition, the dataset was randomly halved into a test set and a training set $\mathcal{D}$. From the training set, a small ($\sim$0.2\%) random subset retain their corresponding ground-truth labels. These data form the initialised labelled dataset $\mathcal{D}_l$. The remaining majority of data from $\mathcal{D}$ have their ground-truth labels hidden, forming the unlabelled dataset $\mathcal{D}_u$.

\begin{figure}[ht!]
		\begin{subfigure}{.5\textwidth}
			\centering
			\scalebox{0.4}{
				\includegraphics{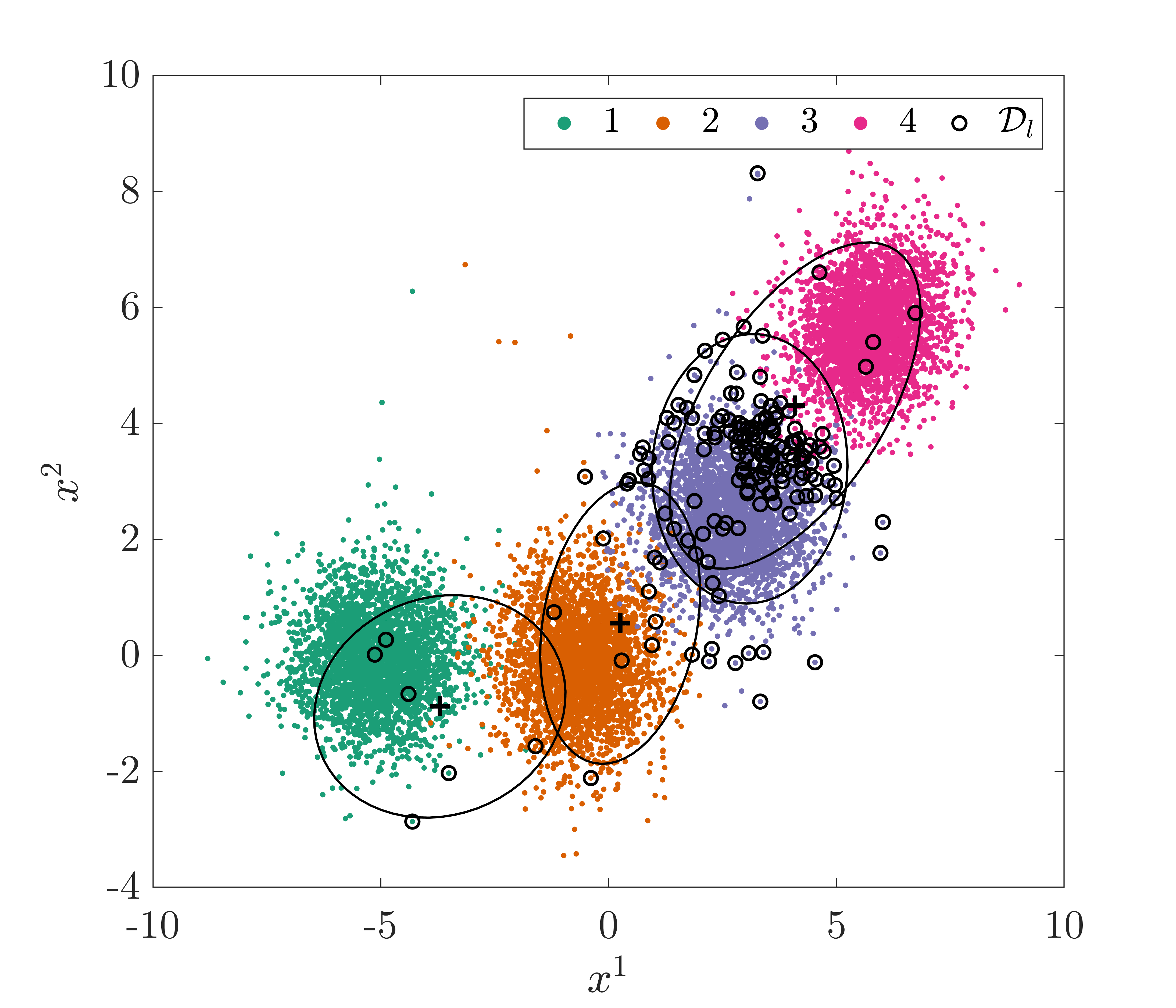}
			}
			\caption{ }
			\label{fig:finalScatter}
		\end{subfigure}
		\begin{subfigure}{.5\textwidth}
			\centering
			\scalebox{0.4}{
				\includegraphics{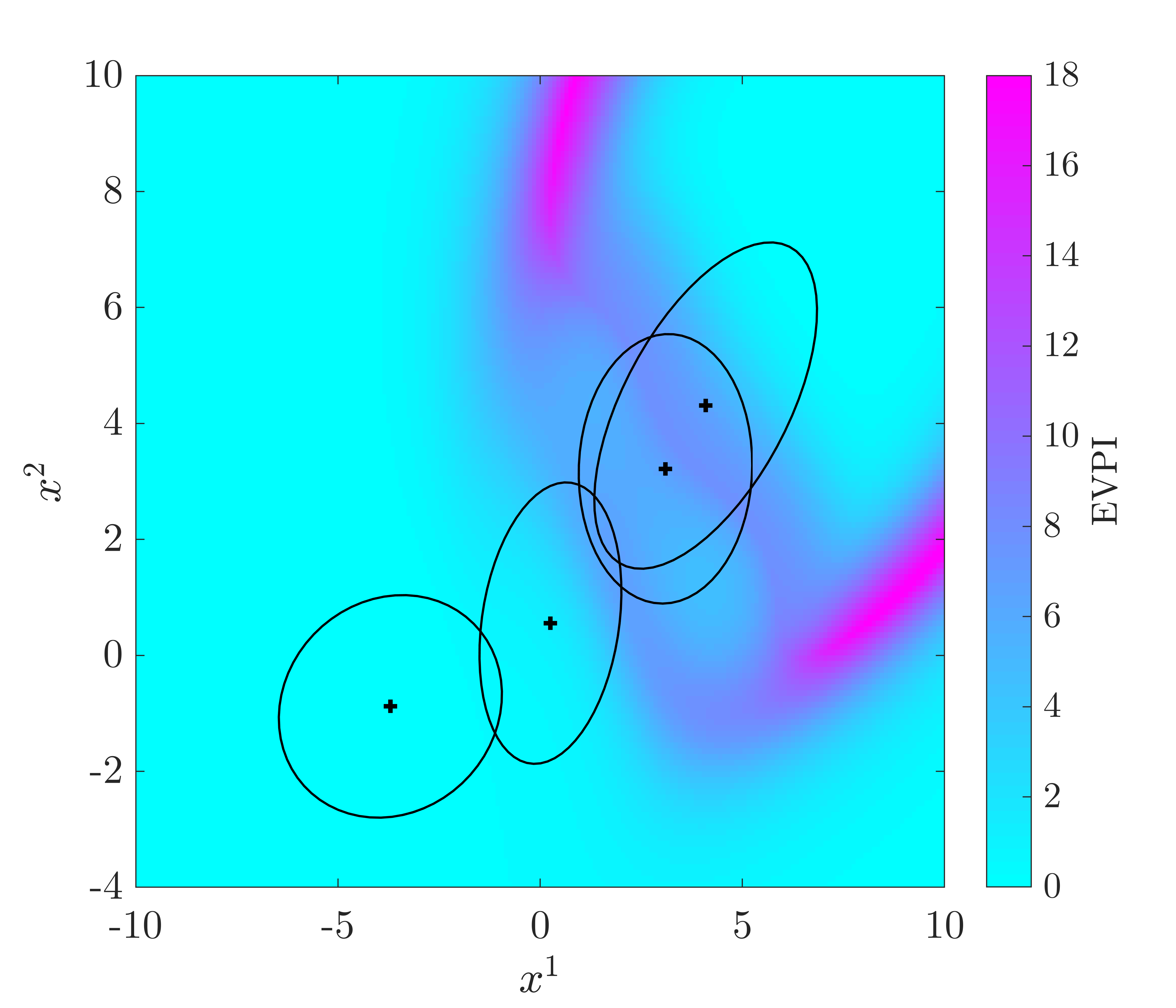}
			}
			\caption{ }
			\label{fig:finalVOI}
		\end{subfigure}
		\caption{A statistical classifier $p(y_t,\mathbf{x}_t,\bm{\Theta})$ following risk-based active learning; \textit{maximum a posteriori} (MAP) estimates of the mean (+) and covariance (ellipses represent 2$\sigma$) are shown. (a) shows the final model overlaid onto the data with labelled data $\mathcal{D}_l$ encircled and (b) shows the resulting EVPI over the feature space.}
		\label{fig:finalModel}
	\end{figure}

	Figure \ref{fig:finalScatter} shows a GMM from one of the 1000 repetitions after the risk-based active-learning process. From Figure \ref{fig:finalScatter}, one can see that during the active-learning process, querying is concentrated in highly-localised regions of the feature space. Specifically, regions between the clusters for Class 3 (moderate damage) and Class 4 (severe damage) have been queried preferentially. It is clear from Figure \ref{fig:finalScatter} that the subset of data $\mathcal{D}_l$ is not representative of the underlying distribution indicating that sampling bias is present. Nonetheless, the queried data have been somewhat successful in learning a decision boundary - as can be deduced by observing the region of high EVPI between the means of clusters for Class 3 and Class 4 in Figure \ref{fig:finalVOI}.
	
	The EM approach to semi-supervised learning was incorporated into the risk-based active learning process and applied to the case study. Once again, 1000 repetitions were conducted, each with randomly selected training and test datasets and with $\mathcal{D}_l$ randomly initialised as a small subset of the training data.
	
\begin{figure}[ht!]
		\begin{subfigure}{.5\textwidth}
			\centering
			\scalebox{0.4}{
				\includegraphics{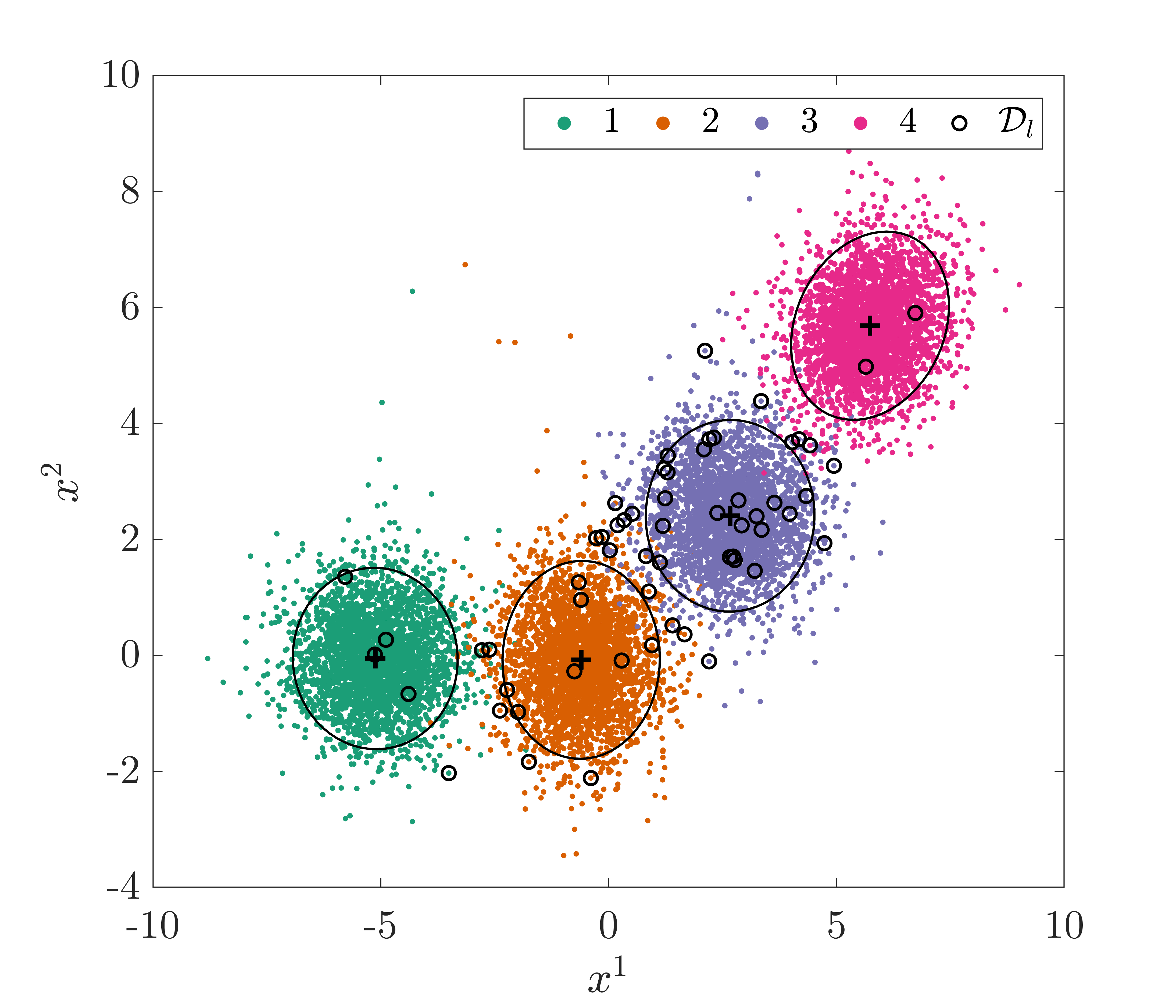}
			}
			\caption{ }
			\label{fig:finalScatter_EM}
		\end{subfigure}
		\begin{subfigure}{.5\textwidth}
			\centering
			\scalebox{0.4}{
				\includegraphics{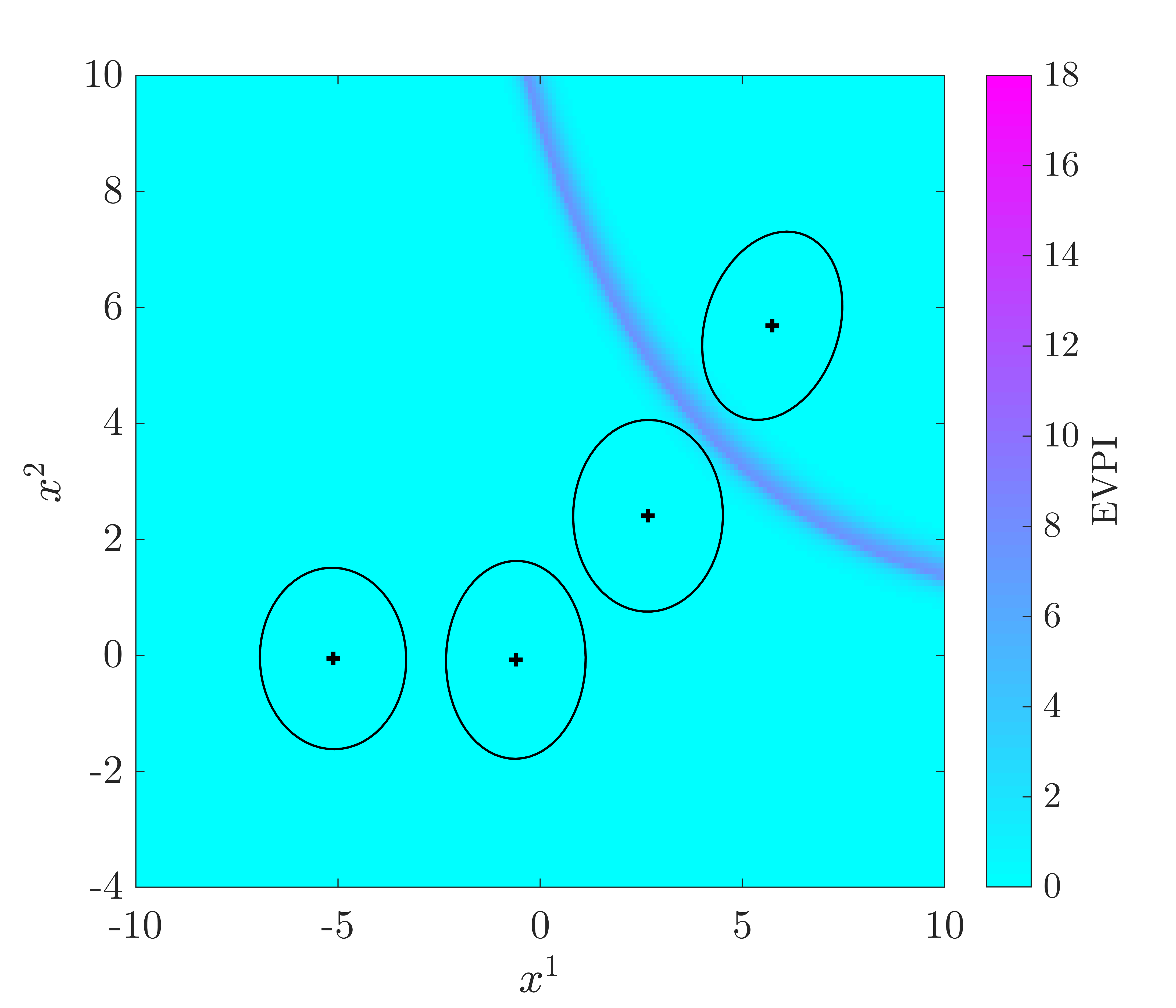}
			}
			\caption{ }
			\label{fig:finalVOI_EM}
		\end{subfigure}
		\caption{A statistical classifier $p(y_t,\mathbf{x}_t,\bm{\Theta})$ following risk-based active learning with semi-supervised learning via EM; \textit{maximum a posteriori} (MAP) estimates of the mean (+) and covariance (ellipses represent 2$\sigma$) are shown. (a) shows the final model overlaid onto the data with labelled data $\mathcal{D}_l$ encircled and (b) shows the resulting EVPI over the feature space.}
		\label{fig:finalModel_em}
	\end{figure}

	Figure \ref{fig:finalModel_em} shows a GMM for one of the 1000 runs after the risk-based active learning process incorporating EM.
	
	It can be seen from Figure \ref{fig:finalScatter_EM} that, similar to the GMM without semi-supervised learning, risk-based active learning with semi-supervised learning results in labels being obtained for localised regions of the feature space, with Class 3 (moderate damage) being preferentially queried. Nonetheless, this figure shows that the clusters learned in a semi-supervised manner fit the data very well. Furthermore, examination of Figure \ref{fig:finalVOI_EM} reveals that the resulting EVPI distributions over the feature spaces distinctly differ from that in Figure \ref{fig:finalVOI}. For the EM approach, the introduction of semi-supervised learning into the risk-based active-learning process has enabled the inference of a well-defined decision-boundary indicated by the band of high EVPI separating the $2\sigma$ clusters for Class 3 and Class 4.
	
\begin{figure}[ht!]
		\centering
		\scalebox{0.4}{
			\includegraphics{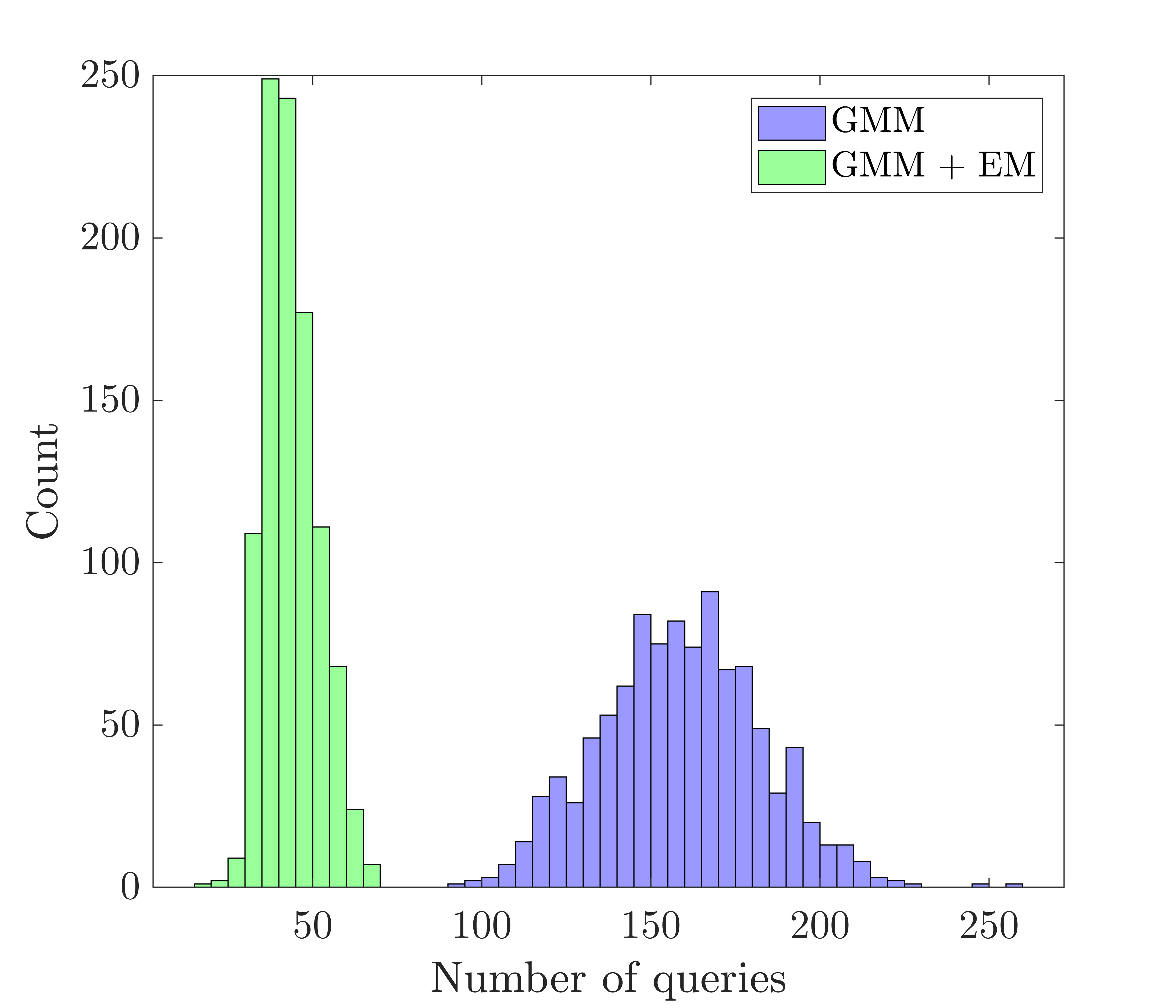}
		}
		\caption{Histograms showing the distribution of the number of queries from 1000 runs of the risk-based active learning of (i) a GMM (blue) and (ii) a GMM semi-supervised via expectation-maximisation (green).}
		\label{fig:hist_ss}
	\end{figure}

\begin{figure}[ht!]
		\centering
		\scalebox{0.4}{
			\includegraphics{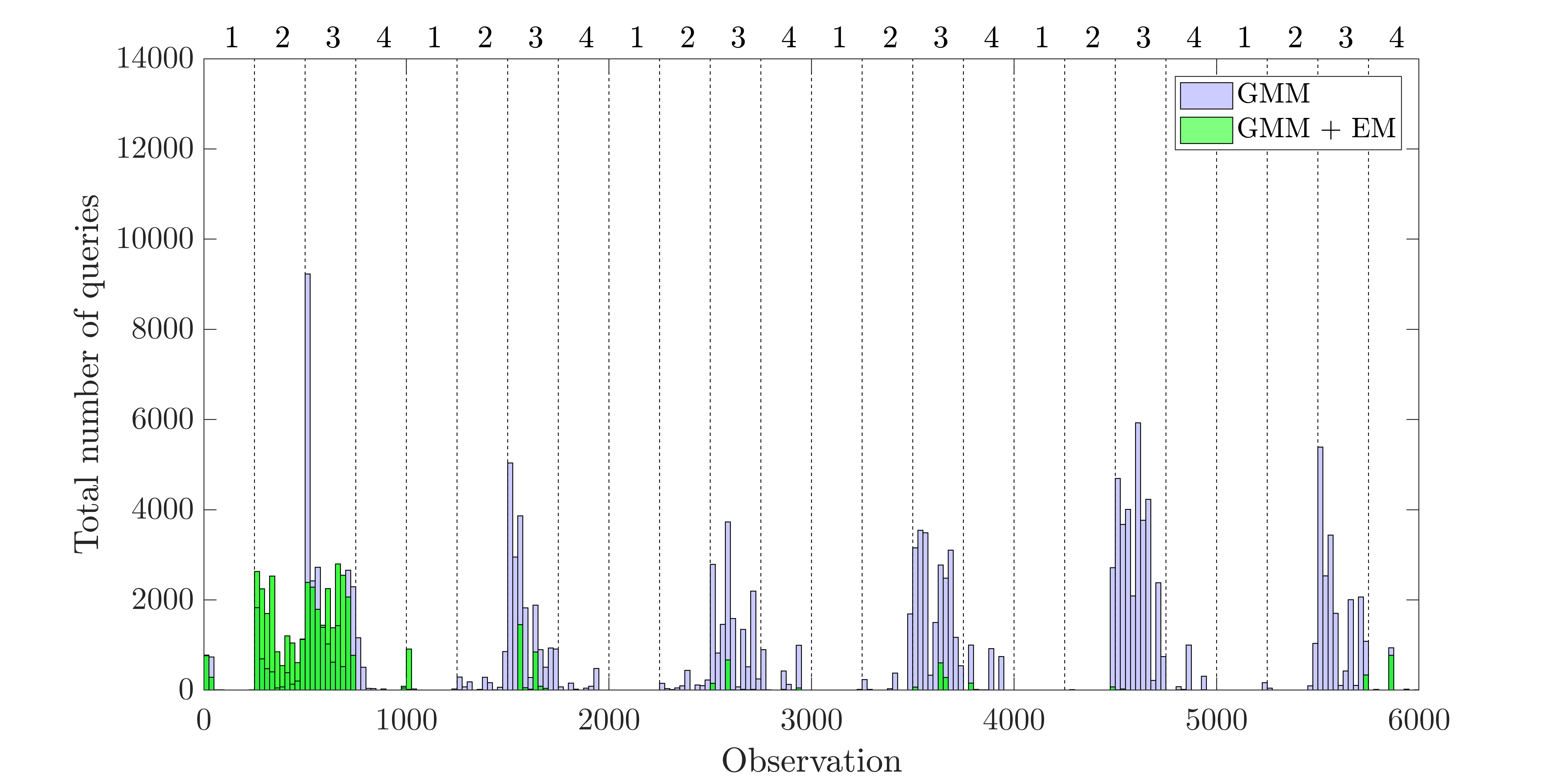}
		}
		\caption{Histograms showing the distribution of the number of queries for each observation in $\mathcal{D}_u$ from 1000 runs of risk-based active learning for (i) a GMM (blue) and (ii) a GMM with EM (green). The average location of classes within $\mathcal{D}_u$ are numbered on the upper horizontal axis and transitions are denoted as a dashed line.}
		\label{fig:all_queries_ss}
	\end{figure}

	Figure \ref{fig:hist_ss} shows how the number of queries varies between each approach to risk-based active learning. It is immediately obvious from Figure \ref{fig:hist_ss}, that incorporating semi-supervised learning reduces the number of queries made substantially. The significance of this result becomes most apparent if one recalls that, in the posed SHM decision context, the number of queries can be mapped directly onto inspection expenditure. This result is to be expected as, when employing semi-supervised learning, each time a query is made, additional information from the unlabelled dataset is utilised to define a class. This allows clusters to become well-defined  more quickly, reducing the area of high-EVPI regions (as is visible in Figure \ref{fig:finalModel_em}), meaning fewer queries are made later in the dataset. This result is evident from Figure \ref{fig:all_queries_ss}.
	
	Figure \ref{fig:all_queries_ss} compares the total number of queries for each index in $\mathcal{D}_u$ over the 1000 repetitions of risk-based active learning conducted with a GMM, and a GMM with EM. It can be seen from Figure \ref{fig:all_queries_ss}, that the incorporation of semi-supervised learning into the risk-based active approach results in relatively more queries being obtained during the first occurrence of each class, with relatively fewer queries being made at later occurrences. It can be seen that Class 3 is heavily investigated at each occurrence when semi-supervised methods are not employed. This phenomenon is to be expected when one considers the differences between the high-EVPI regions shown in Figure \ref{fig:finalVOI} and Figure \ref{fig:finalVOI_EM}; as previously discussed, semi-supervised learning results in a well-defined decision boundary, thereby reducing the likelihood that data will have high value of information.

\begin{figure}[ht!]
		\begin{subfigure}{.5\textwidth}
			\centering
			\scalebox{0.4}{
				\includegraphics{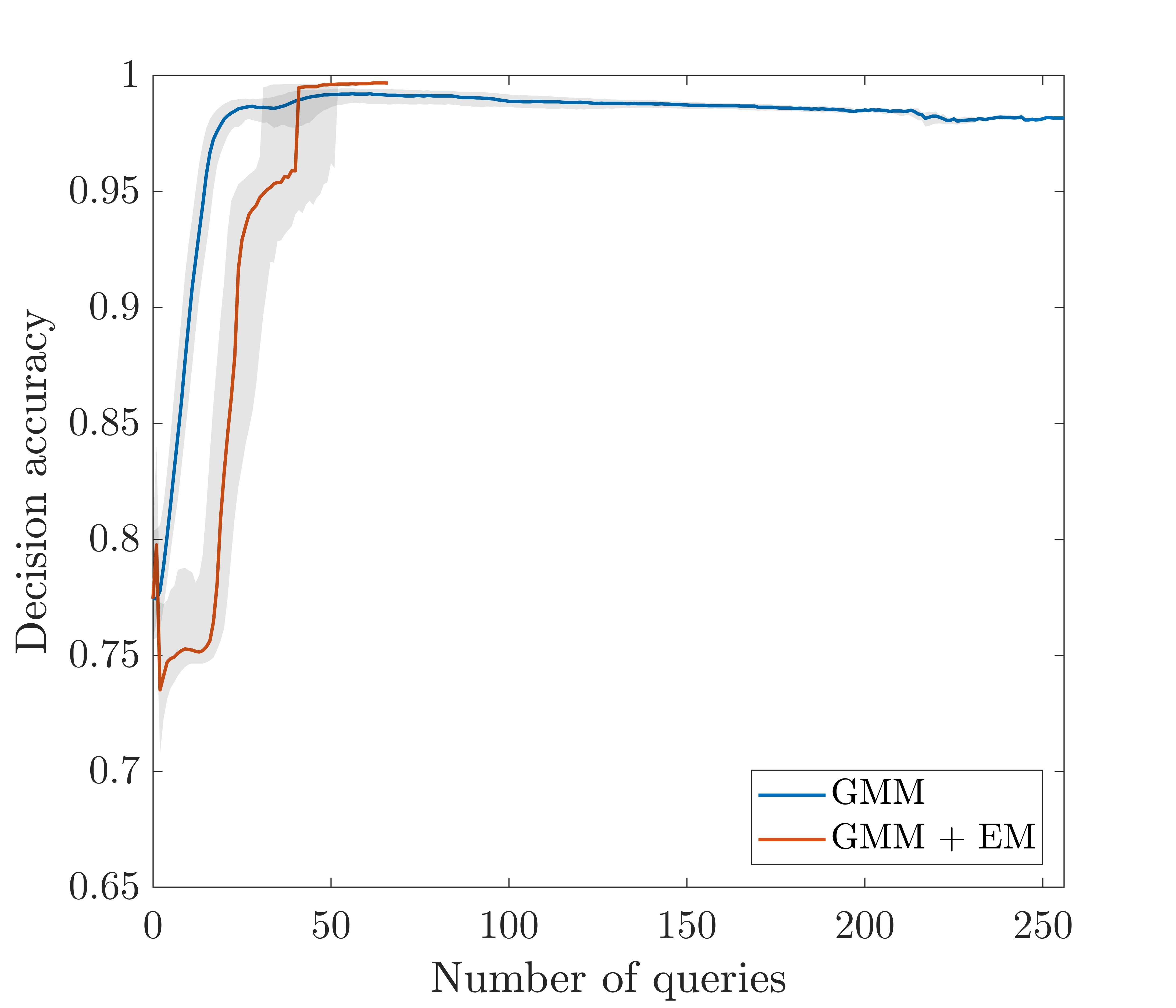}
			}
			\caption{ }
			\label{fig:dacc_ss} 
		\end{subfigure}
		\begin{subfigure}{.5\textwidth}
			\centering
			\scalebox{0.4}{
				\includegraphics{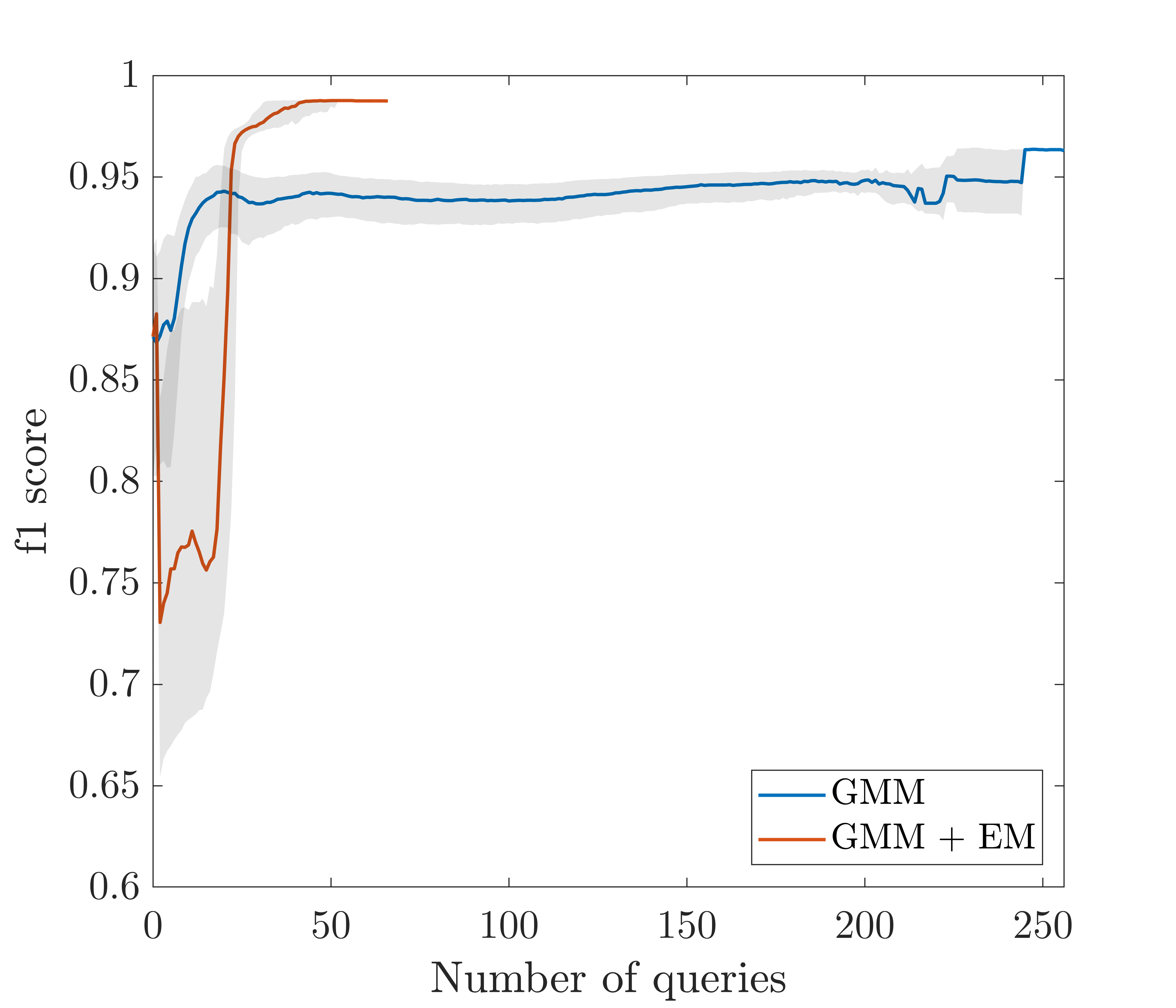}
			}
			\caption{ }
			\label{fig:f1score_ss}
		\end{subfigure}
		\caption{Variation in median (a) decision accuracy and (b) $f_{1}$ score with number of label queries for an agent utilising a GMM learning via risk-based active learning (i) without semi-supervised updates and (ii) with semi-supervised updates via EM. Shaded area shows the interquartile range.}
		\label{fig:performance_ss}
	\end{figure}

	Figure \ref{fig:performance_ss} provides a comparison of the median decision accuracies and $f_1$-scores throughout the querying process between risk-based active learning, with and without semi-supervision.
	
	Here, the decision-making performance achieved via each classifier is assessed via a quantity termed `decision accuracy'. The decision accuracy is a comparison between the actions selected by an agent using the classifier being evaluated, and the optimal (correct) actions selected by an agent in possession of perfect information. For details of this performance measure, the reader is directed to \cite{Hughes2021}.
	
	From Figure \ref{fig:dacc_ss}, one may be inclined to deduce that the introduction of semi-supervised learning has been detrimental to the performance of risk-based active learning as the introduction of EM appears to delay the observed increase in decision accuracy. However, when considered alongside Figure \ref{fig:all_queries_ss}, one can realise that, because semi-supervised learning results in increased querying early in the dataset, decision accuracy over the whole dataset is improved. Furthermore, a decline in decision performance is not observed for the algorithm incorporating semi-supervised learning; this result is because of the reduction in sampling bias obtained via the inclusion of unlabelled data. Figure \ref{fig:f1score_ss} shows the $f_1$-score classification performance, these results provide further indication that the models learned via semi-supervised risk-based active learning better represent the underlying distribution of data and that the detrimental effects of sampling bias have been reduced.
	
	In summary, the case study has shown that, for some applications, semi-supervised learning provides a suitable approach to reducing the effects of sampling bias. The generative distributions obtained via these semi-supervised risk-based active-learning approaches better fit the underlying data distributions, whilst also establishing well-defined decision boundaries in a cost-effective manner.
	
	\section{Conclusions}
	
	To conclude, it has been shown that risk-based active learning of a statistical classifier can improve decision-making performance over the course of an SHM campaign. However, because of issues relating to sampling bias, degradation in performance is observed in the latter stages of the learning process. To rectify this issue, a novel approach to risk-based active learning was presented, in which semi-supervised learning, by means of the EM algorithm, was demonstrated using a numerical case study. The results of the case study showed that decision-making performance can be improved, and sampling bias is counteracted as a well-defined decision boundary is inferred with the additional information provided by the unlabelled data. Furthermore, the EM approach to risk-based active learning significantly reduced the number of queries made, suggesting that resource expenditure during monitoring campaigns can be reduced via careful design of statistical classifiers.


\section*{Acknowledgments}

The authors would like to acknowledge the support of the UK EPSRC via the Programme Grant EP/R006768/1. KW would also like to acknowledge support via the EPSRC Established Career Fellowship EP/R003625/1. LAB was supported by Wave 1 of The UKRI Strategic Priorities Fund under the EPSRC Grant EP/W006022/1, particularly the \textit{Ecosystems of Digital Twins} theme within that grant and The Alan Turing Institute. The authors would like to thank Dr.\ Robert Barthorpe of the University of Sheffield for providing valuable discussion.

	
	\bibliographystyle{splncs04}
	\bibliography{IWSHM2021_SS}

\end{document}